\documentclass[a4paper,conference]{IEEEtran}
\IEEEoverridecommandlockouts
\usepackage{cite}
\usepackage{diagbox}
\usepackage{graphicx}
\usepackage{subfigure}
 \graphicspath{{Image/}}
\usepackage{multirow}
\ifCLASSINFOpdf
\else
\DeclareGraphicsExtensions{.eps}
\fi
\usepackage[backref]{hyperref}
\usepackage{amsmath}
\usepackage{cleveref} 
\usepackage[table,xcdraw]{xcolor}
\usepackage{amssymb}
\usepackage{bbding}
\usepackage{algorithmic}
\usepackage{algorithm}
\usepackage{color}

\hypersetup{hidelinks,
	colorlinks=true,
	allcolors=black,
	pdfstartview=Fit,
	breaklinks=true}

\usepackage{url}

\begin{document}

\title{A Novel Splitting Criterion Inspired by Geometric Mean Metric Learning  for Decision Tree}

\author{\IEEEauthorblockN{Dan Li, Songcan Chen*\thanks{* Songcan Chen is the corresponding author.}}
\IEEEauthorblockA{College of Computer Science and Technology\\
	Nanjing University of Aeronautics and Astronautics\\
	Nanjing, China \\
Email: \{lidanjk, s.chen\}@nuaa.edu.cn}}

\maketitle

\begin{abstract}
Decision tree (DT) attracts persistent research attention due to its impressive empirical performance and interpretability in numerous applications. However, the growth of traditional yet widely-used univariate decision trees (UDTs) is quite time-consuming as they need to traverse all the features to find the splitting value with the maximal reduction of the impurity at each internal node. In this paper, we newly design a splitting criterion to speed up the growth. The criterion is induced from Geometric Mean Metric Learning (GMML) and then optimized under its diagonalized metric matrix constraint, consequently, a closed-form rank of feature discriminant abilities can at once be obtained and the top 1 feature at each node used to grow an intent DT (called as dGMML-DT, where d is an abbreviation for diagonalization). We evaluated the performance of the proposed methods and their corresponding ensembles on benchmark datasets. The experiment shows that dGMML-DT achieves comparable or better classification results more efficiently than the UDTs with 10x average speedup. Furthermore, dGMML-DT can straightforwardly be extended to its multivariable counterpart (dGMML-MDT) without needing laborious operations. 
\end{abstract}


\IEEEpeerreviewmaketitle

\section{Introduction}

Decision tree (DT) has a long history in the fields of pattern recognition, dating back to early versions like CHAID\cite{kass1980exploratory}, ID3, CART\cite{breiman2017classification} and C4.5\cite{quinlan1993c} in the 1980s and 1990s. DT is an interpretable model that can be explained by the if-then rule and commonly used in intrusion detection\cite{ahmim2019novel}, finance\cite{machado2019lightgbm}, health care\cite{shouman2011using} and online purchase behaviour analysis\cite{cho2002personalized} that require interpretability to justify the output of models.

DT is a greedy algorithm that iteratively divides the dataset into disjoint partitions using some splitting criteria, where each internal node is a boolean test on a feature, each branch represents the output of the test, and finally, each leaf node represents a category or response variable.
However, the splitting criteria in DTs are based on optimizing some splitting criteria over all possible splits on all features. In addition, exhaustive search for the optimal axis-parallel hyperplane is NP-hard\cite{warnow2018supertree}\cite{kvrivanek1986np}\cite{laurent1976constructing}\cite{bertsimas2017optimal}\cite{carreira2018alternating}. As a result, a huge number of features will cause curse of dimensionality and a significant increase in computational cost.

Broadly speaking, tree construction primarily spans two themes, univariate decision trees (UDTs) and multivariate decision tree (MDTs), depending on whether the splitting hyperplane of the tree is axis-parallel or oblique. For UDTs, numerous splitting criteria have been proposed. Kass\cite{kass1980exploratory} adopts a testing procedure based on Pearson’s chi-squared statistic to find the best splits. C4.5\cite{quinlan1993c} has been proposed to correct the bias of ID3 trees by incorporating Information Gain Ratio. Breiman and Friedman\cite{breiman2017classification} introduce Classification and Regression Trees (CART) which provides Gini Index and towing criterion as impurity measurement. More recently, Akash et al.\cite{akash2019inter} propose a splitting criterion called Inter-node Hellinger Distance (iHD), which measures the distance between parent and children nodes using Hellinger Distance. The splitting criteria of the above described and our proposal algorithms are compared in Table \ref{sc} below.

All of the aforementioned methods follow the idea of an exhaustive search for the best features and thresholds using corresponding splitting criteria. The main difference among these methods is the measurement of the goodness of a split. As we mentioned above, this kind of exhaustive searches will cause curse of dimensionality\cite{liu2016sparse} at the presence of tremendous features. Most existing methods accelerate tree construction by designing a fast tree-growing algorithm, parallelization\cite{chen2015xgboost}, or data partitioning. Many variants of UDTs have been developed to find alternatives to time-consuming exhaustive searches, such as Very Fast C4.5 (VFC4.5)\cite{cherfi2018very}, one-level decision trees\cite{iba1992induction} and fast decision tree\cite{purdilua2014fast}. However, these methods are different from ours as we can rapidly obtain discriminative ranking of all features at once. Furthermore, the emergence of multivariate decision trees, like OC1\cite{murthy1994system}, Fisher’s decision tree\cite{lopez2013fisher}, multisurface proximal Support Vector Machine based decision tree (MPSVM-based DT)\cite{zhang2014oblique} and Heterogeneous oblique random forest\cite{katuwal2020heterogeneous}  has mitigated this problem to some extent. Even though MDTs are able to avoid exhaustive searches and simplify the classification hyperplane, they obviously compromise the interpretability of DTs.

To cast off such a dilemma, we are motivated to propose a novel UDT, named dGMML-DT, which could rapidly find the best splitting feature via a strictly convex objective function instead of exhaustive searches. Specifically, the closed-form solution of the objective function corresponds to the importance weight of each feature while the feature with the greatest importance weight is used to divide each  internal node of a tree into two disjoint partitions. Notably, dGMML-DT takes into account the geometric distribution of the data. In this way, it simplifies tree structure and accelerates the building process. Therefore, this
method has the advantage of efficiently handling large datasets.  

\begin{table}[t]
\centering  
\caption{Splitting Criteria of DTs}  
\label{sc}  
\begin{tabular}{llc}
\hline
Algorithm                        & Splitting Criteria                             & Exhaustive Search    \\ \hline
\cellcolor[HTML]{F2F2F2}ID3      & \cellcolor[HTML]{F2F2F2}Information Gain       &  \cellcolor[HTML]{F2F2F2}\Checkmark                  \\
CART                             & Gini Index/Towing Criteria                     & \Checkmark                    \\
\cellcolor[HTML]{F2F2F2}C4.5     & \cellcolor[HTML]{F2F2F2}Information Gain Ratio & \cellcolor[HTML]{F2F2F2}\Checkmark                   \\ 
iHD                              & Inter-node Hellinger Distance                  & \Checkmark                  \\
\cellcolor[HTML]{F2F2F2}dGMML-DT & \cellcolor[HTML]{F2F2F2}Feature Weight         & \cellcolor[HTML]{F2F2F2}\XSolidBrush    \\ \hline        
\end{tabular}
\end{table}

In addition, our algorithm is relatively straightforward and flexible, so that it can construct two types of trees (axis-parallel and oblique) at once. dGMML-DT can be expanded into its multivariable counterpart (dGMML-MDT) easily. It is worthwhile noting that though the proposed dGMML-DT is only a simple attempt, the method is generalizable enough to be embedded into other variations of DTs.
In this paper, we also consider the random forest\cite{breiman2001random} of dGMML-DT, namely dGMML-RF, to reduce variance via Bagging\cite{breiman1996bagging} and Random-Subspaces\cite{ho1998random}. We will skip over the technical details on random forest, we refer readers to \cite{2015Decision} for more details.

The main contributions of the proposed method are three-fold:\begin{enumerate}
\item We propose a novel splitting criterion by cooperating with Geometric Mean Metric Learning (GMML) under the diagonalization constraint of metric matrix to obtain the discriminant rank of features at each internal node. 
\item dGMML-DT is generalizable to other variants of DTs and significantly narrows the search space so as to reduce the inference time. 
\item We evaluate the performance of the proposed method and its corresponding ensemble on benchmark datasets taken from the UCI repository and other real-world datasets. They achieve comparable or better accuracy than other DTs more effectively and efficiently.
\end{enumerate}

The remainder of this paper is organized as follows. We represent a brief review of typical splitting criteria and the latest related work in Section \ref{related work}. Our proposed splitting criterion is delineated in Section \ref{proposed method}.  In Section \ref{Experimental results}, we present the experimental study where new methods are compared with typical splitting criteria and state-of-art classifiers. Finally, Section \ref{conclusion} concludes the paper.

\section{Related Work}\label{related work}
The splitting criterion is an important issue in the induction of DTs. The feature with best score which reduces the average uncertainty mostly will be chosen as the splitting feature for a node. In this section, we will discuss several splitting criteria briefly .

To simplify the description of the related works, suppose for node $t$, the label $y$ takes values of $1,2,\ldots,c$, and $p_{t i}\ (i=1,2,\ldots,c)$ represents the proportion of class $i$ in node $t$. 
\subsection{Information Gain}
Information entropy is a measure of the information uncertainty and can be used to measure the impurity of a node. The information entropy of node $t$ is:
$$\operatorname{H}(t)=-\sum_{i=1}^{c} p_{t i} \log _{2} p_{t i}.$$
When all the samples of $t$ belong to the same category, the information entropy is zero and the impurity of the node is minimal. Based on information entropy, the reduction of information entropy after node splitting can be calculated and called information gain:
$${ IG(t,\theta) }=\operatorname{H}\left(t\right)- \frac{n_{l}}{n} \operatorname{H}\left(l\right)-\frac{n_{r}}{n} \operatorname{H}\left(r\right),$$
where $l$ and $r$ denote two child nodes of $t$, $n_l$ and $n_r$ are the number of samples in $l$ and $r$ respectively. $\theta$ represents the splitting feature and threshold of node $t$. The parameter $\theta$ with maximum information gain will be selected to split node $t$.
\subsection{Gini Index}
Gini Index reflects the degree of homogeneity of different categories in a node. The Gini Index of node $t$ is:
$$\operatorname{G}(t)=1-\sum_{i=1}^{c}p_{t i}^{2}.$$
The reduction of impurity (IR) after splitting $t$ is:
$${ IR(t,\theta) }=\operatorname{G}\left(t\right)- \frac{n_{l}}{n} \operatorname{G}\left(l\right)-\frac{n_{r}}{n} \operatorname{G}\left(r\right).$$

\subsection{Misclassification Error}
When the label of the current node is assigned according to the majority class, the proportion of misclassified samples can be used to measure the impurity of node $t$:
$$Error(t)=1-\max _{1 \leq i \leq c} p_{t i}.$$
\subsection{Inter-node Hellinger Distance }
Akash et al.\cite{akash2019inter} proposed a skew insensitive splitting criterion called Inter-node Hellinger Distance (iHD) that uses the squared Hillinger distance ($D_{H}^{2}$) to measure the heterogeneity between the class probability distributions of the parent and child nodes.

Suppose $\mathbf{P}_{t}\left(p_{t 1}, \ldots, p_{t c}\right)$ is the class probability distribution for node $t$. $D_{H}^{2}$ between node $t$ and its left child (similar for right child) $D_{H}^{2}\left(\mathbf{P}_{l} \| \mathbf{P}_{t}\right)$ is:
$$D_{H}^{2}\left(\mathbf{P}_{l} \| \mathbf{P}_{t}\right)=1-\sum_{i=1}^{c} \sqrt{p_{l i} p_{t i}}.$$
Then iHD is defined as
$$\text { iHD }=\frac{n_{l}}{n} D_{H}^{2}\left(\mathbf{P}_{l} \| \mathbf{P}_{t}\right)+\frac{n_{r}}{n}D_{H}^{2}\left(\mathbf{P}_{r} \| \mathbf{P}_{t}\right).$$
The feature with maximum iHD will be used to split node $t$.

\section{Proposed Method} \label{proposed method}
Aforementioned splitting criteria of UDTs rely on cumbersome exhaustive searches. However, inspired by GMML, we consider whether the most discriminative features can be rapidly selected by an optimizable discriminatory criterion. The core of dGMML-DT is to cooperate with GMML under the diagonalization constraint of metric matrix to obtain the discriminant rank of features by optimizing a strictly convex objective function. Before presenting our method, we first revisit preliminaries on tree building process.
\subsection{Tree Building Process}
We focus on binary classification problems, it is straightforward to scale our method to multi-class tasks. Without loss of generality, we assume that input values are continuous. We denote training data as $X=\left\{\left(x_{1}, y_{1}\right), \ldots,\left(x_{n}, y_{n}\right)\right\} \subseteq \mathbb{R}^{d} \times\{-1,+1\}$ , where $x_{i}=\left(x_{i 1}, x_{i 2}, \ldots, x_{i d}\right)$ for $i=1,2,\ldots, n$ are input vectors, $d$ is the feature dimension of $x_i$, and $y_{i} \in\{-1,+1\}$ are labels. 

 Define $X^t$ as the data at $t$-th node and then separate $X^t$ into two partitions or children denoted by $l$ and $r$ respectively. Suppose the best splitting feature and threshold are $j^*$ and $b$ respectively. The partition implement is shown below: 
  \begin{equation} \label{1}
 \text { partition } 1: l=\left\{\left(x_{i}, y_{i}\right) \in X^{l} \text {, s.t. } x_{i j^{*}} \leq b\right\},
   \end{equation}
  \begin{equation} \label{2}
  \text { partition } 2: r=\left\{\left(x_{i}, y_{i}\right) \in X^{r}, \text { s.t. } x_{i j^{*}}>b\right\}.
  \end{equation}

 In UDTs, the best selected split parameter $ \theta^*=(j^*,b)$ of node $t$ is the one with maximum reduction of impurity which can be formulated as follows:
 \begin{equation} \label{3}
 \theta^{*}={\arg \max_\theta }\ IR(t , \theta).
 \end{equation}
 
 The DT recursively divides the data into disjoint partitions using equations (\ref{1}), (\ref{2}) and (\ref{3}) until the node becomes pure or some stop criteria are met. 

\subsection{Splitting Criterion}
 For simplicity, in our method, we maximize the distance between class means to make sure points from different classes are far away from each other, while minimizing the variance within each class to ensure similar points are as close as possible.

The objective is to find a diagonal weight matrix $A=diag(w_1,\ldots,w_j,\ldots,w_d)$, where the $j$-th element of $A$ denotes the weight of the feature $j$. 
For brevity, we define within-class covariance (similarity) matrix $S_w$  and between-class covariance (dissimilarity) matrix $S_b$ as follows:
\begin{equation}
   S_{w}=\sum_{c=1}^{2} \sum_{x_{i} \in C}\left(x_{i}-m_{c}\right)\left(x_{i}-m_{c}\right)^{T},
\end{equation}
\begin{equation}
    S_{b}=\left(m_{1}-m_{2}\right)\left(m_{1}-m_{2}\right)^{T},
\end{equation}
where $c$ is the class number, $m_c$ is the mean of class $C$. Hence, $m_1$ and $m_2$ represent the mean of positive and negative class respectively. Moreover, as in \cite{zadeh2016geometric}, minimizing
$d_{A}\left(x, x^{\prime}\right)=\left(x-x^{\prime}\right)^{T} A\left(x-x^{\prime}\right)$ is equivalent to maximizing $d_{A^{-1}}\left(x, x^{\prime}\right)$, where $x$, $x^{'}$ are samples and $A$ is a $d \times d$ real symmetric positive definite matrix. Inspired by this formulation, the objective function of our splitting criterion at each internal node $t$ is:
\begin{equation} \label{ob}
\min _{A \succ \mathbf{0}} f(A)=\operatorname{tr}\left(A S_{w}\right)+\operatorname{tr}\left(A^{-1} S_{b}\right),
\end{equation}
where $A\succ \mathbf{0}$ represents $A$ is a positive definite matrix.  The formula (\ref{ob}) can be further rewritten as:
\begin{equation}
\begin{split}
    f\left(w_{1}, \ldots, w_{j}, \ldots, w_{d}\right)=\sum_{j=1}^{d} \sum_{c=1}^{2} \sum_{x_{i}\in C} w_{j}\left(x_{i j}-m_{c j}\right)^{2}+ \\
    \sum\limits_{j=1}^{d} \frac{1}{w_{j}}\left(m_{1 j}-m_{2 j}\right)^{2},
\end{split}
\end{equation}
where $x_{ij}$ and $m_{cj}$ represent the $j$-th feature of the sample $x_i$ and class $C$’s mean respectively. The optimization problem of (\ref{ob}) is strictly and geodetically convex. In other words, when the derivative of $f(A)$ is zero, the minimizer we obtain is a global optimal solution. Differentiating with respect to $w_j$, we obtain
\begin{equation} \label{deri}
\begin{split}
\frac{\partial f\left(w_{1}, \ldots, w_{j}, \ldots, w_{d}\right)}{\partial w_{j}}=\sum_{c=1}^{2} \sum_{x_{i} \in C}\left(x_{i j}-m_{c j}\right)^{2}-\\ \frac{1}{w_{j}^{2}}\left(m_{1 j}-m_{2 j}\right)^{2}.
\end{split}
\end{equation}
Setting the derivative to zero to find the optimal solution of (\ref{ob}), we yield
\begin{equation} \label{wj}
    w_{j}=\sqrt{\frac{\left(m_{1 j}-m_{2 j}\right)^{2}}{\sum\limits_{c=1}^{2} \sum\limits_{x_{i} \in C}\left(x_{i j}-m_{c j}\right)^{2}}}.
\end{equation}

The intuition of Equation (\ref{wj}) is obvious: if two class centres are further apart and the distribution within each class is more compact, the feature $j$ is more distinguishable and naturally has a greater importance weight in classification. The optimal splitting feature $j^*$ of the node corresponds to the top 1 feature with the highest weight. $j^*$ is represented as:
\begin{equation}\label{10}
    j^{*}=\arg \max _{j} w_{j}, \quad j=1, \ldots, d
\end{equation}

The next challenge is to find the splitting point $b$. We propose three ways to obtain $b$. Firstly, we choose five feature values closest to each other of the two classes respectively and set the mean of these ten values as the splitting point $b$. However, the number of values is a hyperparameter (default 10) which is set manually according to different datasets. Suppose the mean of feature $j^*$ in positive class is smaller than that in negative class, then the splitting point can be expressed by the following equation:
\begin{equation}\label{mean12}
    b = mean(b_1+b_2),
\end{equation}
where $b_1$ ($b_2$) is the mean of five biggest (smallest)  values of feature $j^*$ in positive (negative) class.
Secondly, we set the median of the $j^*$-th feature as the splitting point $b$ since the median is robust, then $b$ can be represented as: 
\begin{equation} \label{medi}
    b=median(x_{i j^{*}}), \quad i=1, \ldots, n
\end{equation}
Lastly, we set the mean of feature $j^*$ as $b$,
\begin{equation} \label{mean}
    b=mean(x_{i j^{*}}), \quad i=1, \ldots, n
\end{equation}

\begin{algorithm}[t]
    \caption{dGMML-DT induced by GMML}
    \label{alg1} 
    \begin{algorithmic}[1]
    \renewcommand{\algorithmicrequire}{\textbf{Input:}}
    \renewcommand{\algorithmicensure}{\textbf{Output:}}
    \REQUIRE ~~\\
    \emph{X}: training dataset with n samples.\\
    \emph{mtry}: number of candidate features to be evaluated at each internal node.\\
    \emph{minleaf}: minimum number of samples per leaf.\\
    \ENSURE ~~\\ 
    \emph{dGMML-DT}: a decision tree
    \STATE View the node $X^{t}$ as the root node.
    \IF{samples in $X^{t}$ are from the same class C } 
     {\RETURN $X^{t}$ as a leaf node and assign it to class C;}
     \ELSIF {$X^{t}$ contains less than $minleaf$ samples} 
     {\RETURN $X^{t}$ as a leaf node and assign it to the majority class C;}
     \ELSE{
    \STATE Select $mtry$ candidate features uniformly, without replacement.
    \STATE Calculate the importance weight of each candidate feature by Eq.(\ref{ob}).
    \STATE Select the best splitting feature $j^*$ and corresponding splitting point $b$ by Eq.(\ref{10}) and one of Eq.(\ref{mean12})-(\ref{mean}), respectively.
    \STATE 	Split $X^{t}$ into two partitions $X^{l}$ and $X^{r}$ according to Eq.(\ref{1}) and Eq.(\ref{2}).  
    }  
    \ENDIF
    \FOR{each partition}     
       \STATE Repeat lines 1-14 for the partition to grow a subtree.
    \ENDFOR
    \RETURN{dGMML-DT}
    \end{algorithmic}
\end{algorithm}

dGMML-DT grows trees using a divide and conquer partitioning strategy. The best split parameter $ \theta^*=(j^*,b)$ of our splitting criterion at each node $t$ can be chosen by formula (\ref{10}) and one of (\ref{mean12})-(\ref{mean}). The specific process of dGMML-DT is shown in Algorithm \ref{alg1}.

As shown in (\ref{wj}), we can at once get a set of optimized feature importance weights $w$ and thus can directly be used as coefficients to build a linear classifier for growing an oblique decision tree, called dGMML-MDT, since the coefficients in a linear model determine the importance of features\cite{lopez2013fisher}\cite{zhou2021unbiased}, which states our criterion is flexible and so-obtained weights can directly serve the building of both axis-parallel and oblique DTs. Specifically, at each internal node, the individual features are combined into a linear classifier, where $w$ is taken as the coefficient.
\begin{equation}
    p_{i}=w^{t} x_{i}
\end{equation}
The partition implement is shown below:
\begin{equation}
    \text { partition 1: } l=\left\{\left(x_{i}, y_{i}\right) \in \mathcal{X}^{l}, \text { s.t. } p_{i} \leq b\right\},\\
\end{equation}
 \begin{equation}
\text { partition 2: } r=\left\{\left(x_{i}, y_{i}\right) \in \mathcal{X}^{r}, \text { s.t. } p_{i}>b\right\}.
\end{equation}

In addition, our idea is similar to Linear Discriminant Analysis (LDA) to some extent. We have tried to optimize the objective function of LDA to obtain the feature importance weight, but that is too complicated and unattainable to find a closed-form solution like dGMML-DT because LDA objective is nonconvex and generally lacks a closed-form solution for multi-class problems.

\begin{table}[t]
\centering  
\caption{datasets for experiments}  
\label{data}  
\begin{tabular}{lcc}
\cline{1-3}
Datasets                         & \#Samples & \#Features \\
\cline{1-3}
wilt                             & 4839      & 6          \\
kungchi3                         & 123       & 40         \\
knuggetchase3                    & 194       & 40         \\
cloud                           & 108       & 8         \\
spectf                           & 267       & 45         \\
stock                            & 950       & 9          \\
qualitative bankruptcy           & 250       & 7          \\
tokyo1                           & 959       & 45         \\
kc1-top5                         & 145       & 96         \\
diabetes                         & 768       & 9          \\
datatrieve                       & 130       & 9          \\
planning relax                       & 182       & 13       \\
banknote authentication          & 1372      & 5          \\
sonar                    & 208       & 61         \\
heart statlog                    & 270       & 14         \\
blood transfusion center & 748       & 5          \\
breast cancer             & 569       & 32         \\
echocardiogram                   & 132       & 12         \\
fertility                        & 100       & 10         \\
haberman survival                       & 306       & 3         \\
hepatitis                        & 155       & 19         \\
molec biol promoter                        & 106       & 58         \\
parkinsons                       & 197       & 23         \\
planning relax                   & 183       & 13         \\
statlog german credit            & 1000      & 20         \\
\cline{1-3}
\end{tabular}
\end{table}

\section{Experimental results} \label{Experimental results}
In this section, we depict the experimental setup and analyze the performance of the proposed dGMML-DT and other models. We further investigate the inference time of each model. Moreover, we explore the relationship between feature importance weight and impurity. Finally, we compare the performance on three types of splitting points.

\subsection{Experimental Setup}

\begin{table*}[htbp]
\centering  
\caption{Performance of the DTs and RFs based on different split criteria on 25 datasets.}  
\label{accuracy}  
\resizebox{\textwidth}{!}{
\begin{tabular}{|c|cc|cc|cc|cc||c|c|c|c|}
\hline
\multirow{2}{*}{\diagbox{DATASET}{METHOD}}
       & \multicolumn{2}{c|}{C4.5} & \multicolumn{2}{c|}{CART} & \multicolumn{2}{c|}{iHD}  & \multicolumn{2}{c||}{dGMML-DT} & C4.5-RF       & CART-RF       & iHD-RF        & dGMML-RF      \\ 
\cline{2-13}                                & Accu          & Time    & Accu          & Time     & Accu          & Time     & Accu            & Time       & Accu          & Accu          & Accu          & Accu          \\  \hline
wilt                             & 99.5          & 3.702 & 99.8          & 17.826 & \textbf{100}  & 19.621 & \textbf{100}    & \underline{1.21}    & \textbf{100}          & \textbf{100}           & \textbf{100}           & \textbf{100}           \\
kungchi3                         & 83            & 1.673 & \textbf{85.5} & 9.587 & 80.7          & 24.633 & 83.7            & \underline{1.52}    & 86.3          & 86.3          & \textbf{87.2} & 85.7          \\
knuggetchase3                    & 79.1          & 1.785 & 77.9          & 8.74  & 74.1          & 77.97  & \textbf{79.9}   & \underline{0.524}   & 78.5          & 79.6          & 80.6          & \textbf{81.5} \\
cloud                            & \textbf{69.1} & 1.606 & 56.1          & 6.747 & 59.1          & 26.518 & 68.7            &\underline{0.147}   & 69.7          & 60.6          & 64.6          & \textbf{70.2} \\
spectf                           & 67.2          & \underline{1.982} & 75            & 8.929 & 75.2          & 99.14  & \textbf{76.9}   & 2.446   & 79.1          & 79.9          & 80.3          & \textbf{80.9} \\
stock                            & 89.8          & 2.49  & 94.5          & 7.984 & 94.9          & 64.508 & \textbf{95.7}   & \underline{1.508}   & 95.7          & 96.9          & 96.6          & \textbf{97.3} \\
qualitative bankruptcy           & 96.8          & 1.845 & 96.4          & 6.177 & \textbf{99.6} & 6.373 & 98.4            & \underline{0.501}   & 99.2          & 97.6          & \textbf{99.6} & 98.8          \\
tokyo1                           & 87.8          & 2.367 & 91.2          & 10.853 & 90.7          & 178.485 & \textbf{91.3}   & \underline{0.545}   & 92.3          & 93            & 93.6          & \textbf{94.6} \\
kc1-top5                         & 52.9          & 1.616 & \textbf{93}   & 8.894 & 54.8          & 34.435 & \textbf{93}     & \underline{0.422}   & 54.5          & 93.8          & 57.3          & \textbf{94}   \\
diabetes                         & 68.9          & \underline{2.178} & \textbf{72.8} & 9.924 & 71.7          & 114.319 & 72.1            & 5.573   & \textbf{76.9} & 76            & 74.9          & 75.8          \\
datatrieve                       & \textbf{90}   & \underline{1.578} & 86.9          & 8.53  & 87.7          & 8.483 & \textbf{90}     & 4.028   & 90.8          & 90            & \textbf{91.5} & \textbf{91.5} \\
planning relax                   & \textbf{63.1} & 1.684 & 56.5          & 7.472 & 57.2          & 49.771 & 62.6            & \underline{0.764}   & \textbf{72.9} & 65.8          & 64.7          & 70.7          \\
banknote authentication          & 92.5          & 2.906 & \textbf{98.5} & 8.122 & 98.4          & 43.416 & 98              & \underline{0.904}   & \textbf{99.3} & 99.1          & 99            & 99.2          \\
sonar                            & 67            & \underline{1.836} & 71.6          & 8.96  & \textbf{75.9} & 77.619 & 72.7            & 3.031   & 79.4          & \textbf{82.7} & 80.6          & 81.8          \\
heart statlog                    & 76.7          & 1.877 & 78.9          & 7.219 & 77            & 50.497 & \textbf{79.6}   & \underline{1.441}   & 82.6          & 81.1          & 82.2          & \textbf{83}   \\
blood transfusion center & 76.1          & 2.175 & 75.7          & 7.384 & 74            & 91.61  & \textbf{76.9}   & \underline{0.819}   & 78.5          & 76.6          & 76.3          & \textbf{79.8} \\
breast cancer          & 89.6          & 2.139 & 92.3          & 11.125 & 92.6          & 76.656 & \textbf{93}     & \underline{0.638}   & 95.5          & 94.2          & 95.5          & \textbf{96.1} \\
echocardiogram                   & 73.5          & \underline{1.561} & 68            & 6.639 & 80.2          & 15.144 & \textbf{85.5}   & 3.71    & 80.9          & 74.8          & 83.4          & \textbf{86.5} \\
fertility                        & 84            & 1.561 & 85            & 6.273 & 84            & 7.646 & \textbf{87}     & \underline{0.911}   & 86            & 87            & 87            & \textbf{88}   \\
haberman survival                & \textbf{74.2} & 1.764 & 63.8          & 6.862 & 64.6          & 26.274 & 69.2            & \underline{1.341}   & \textbf{73.5} & 73.2          & 71.9          & 72.2          \\
hepatitis                        & 67.5          & 1.675 & 77.7          & 6.894 & 69.8          & 28.626 & \textbf{77.8}   & \underline{1.367}   & 74.2          & 80.2          & 74.8          & \textbf{80.7} \\
molec biol promoter              & 56.3          & 1.606 & 72.9          & 7.897 & \textbf{77.1} & 32.515 & 71.8            & \underline{0.906}   & 73.3          & \textbf{85.1} & 84.1          & 82.8          \\
parkinsons                       & 73.5          & 1.792 & 81.5          & 8.995 & 82.4          & 28.445 & \textbf{83.4}   & \underline{1.386}   & 89            & 92            & 91            & \textbf{92.7} \\
planning                         & 67.3          & \underline{1.697} & 54.8          & 7.272 & 60.4          & 49.674 & \textbf{67.5}   & 3.958   & \textbf{70.4} & 64.3          & 64.9          & \textbf{70.4} \\
statlog german credit            & 70.9          & 2.427 & 71.5          & 9.599 & 71.2          & 321.535 & \textbf{73.5}   & \underline{0.877}   & 72.8          & 74            & 75.7          & \textbf{76.4}  \\ \hline
\end{tabular}
}
\end{table*}

We conduct our experiments on UCI\cite{Dua:2019} and \href{https://www.openml.org/search?type=data}{OpenML} datasets, as summarized in Table \ref{data} (The first 16 datasets are from UCI and the last 9 datasets are from OpenML). These datasets are processed into numerical features. We implement unpruned dGMML-DT, CART, C4.5, iHD and their corresponding ensembles dGMML-RF, CART-RF, C4.5-RF, iHD-RF (RF is the abbreviation of random forest) in Matlab. Moreover, we download the code of the iHD from \href{https://github.com/ZDanielsResearch/HellingerTreesMatlab}{github}. At each non-terminal node, we evaluate $mtry = \sqrt{d}$ number of features, here $d$ is the dimension of the feature set and the $minleaf$ parameter is set to default. We choose formula (\ref{10}) and (\ref{mean12}) to calculate the node splitting feature and point respectively. All the ensemble models consist of 20 base learners. The results are obtained and reported using 10-fold cross-validation on every dataset.
\subsection{Classification Accuracy}
Table \ref{accuracy} summarizes the classification accuracy of each DT on 25 datasets. The highest accuracy is highlighted in bold. From its first four columns, dGMML-DT emerges as the overall winner on 16 datasets. C4.5, CART and iHD beat other models on 4 datasets respectively. It is obvious that dGMML-DT achieves better accuracy compared to other existing DTs, such as C4.5, CART, and iHD, for both small-size and large-size datasets. Similarly, in the last four columns of the table, C4.5-RF, CART-RF, iHD-RF and dGMML-RF achieves the highest accuracy on 6, 3, 4, 17 datasets, respectively. The performance of dGMML-RF is superior on most datasets compared to other RFs evidently.

\begin{figure*}
    \centering
   \subfigure[statlog german credit]{
        \includegraphics[width=0.3\linewidth]{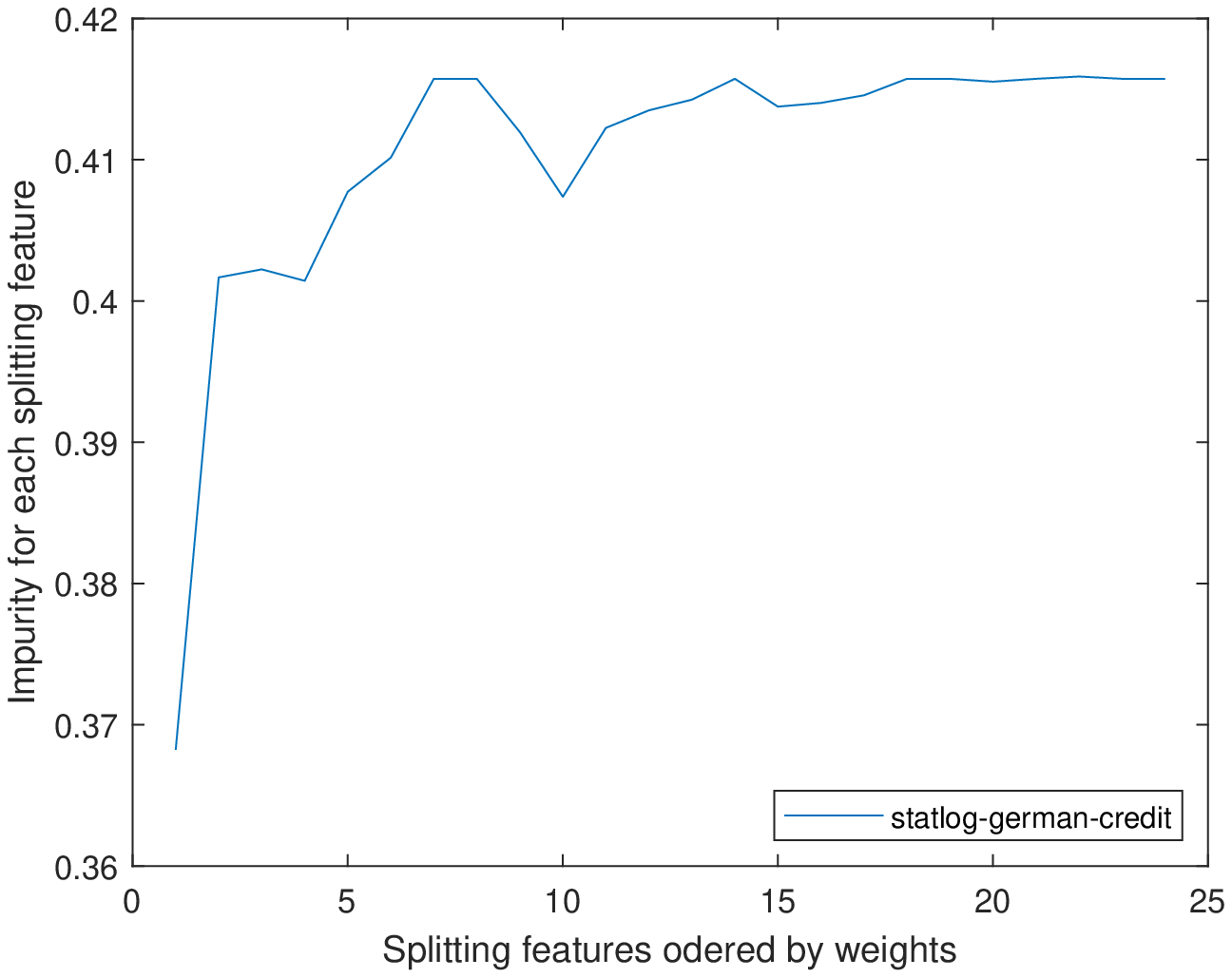}}
     \label{1a} 
     \hfill
   \subfigure[hepatitis]{
        \includegraphics[width=0.3\linewidth]{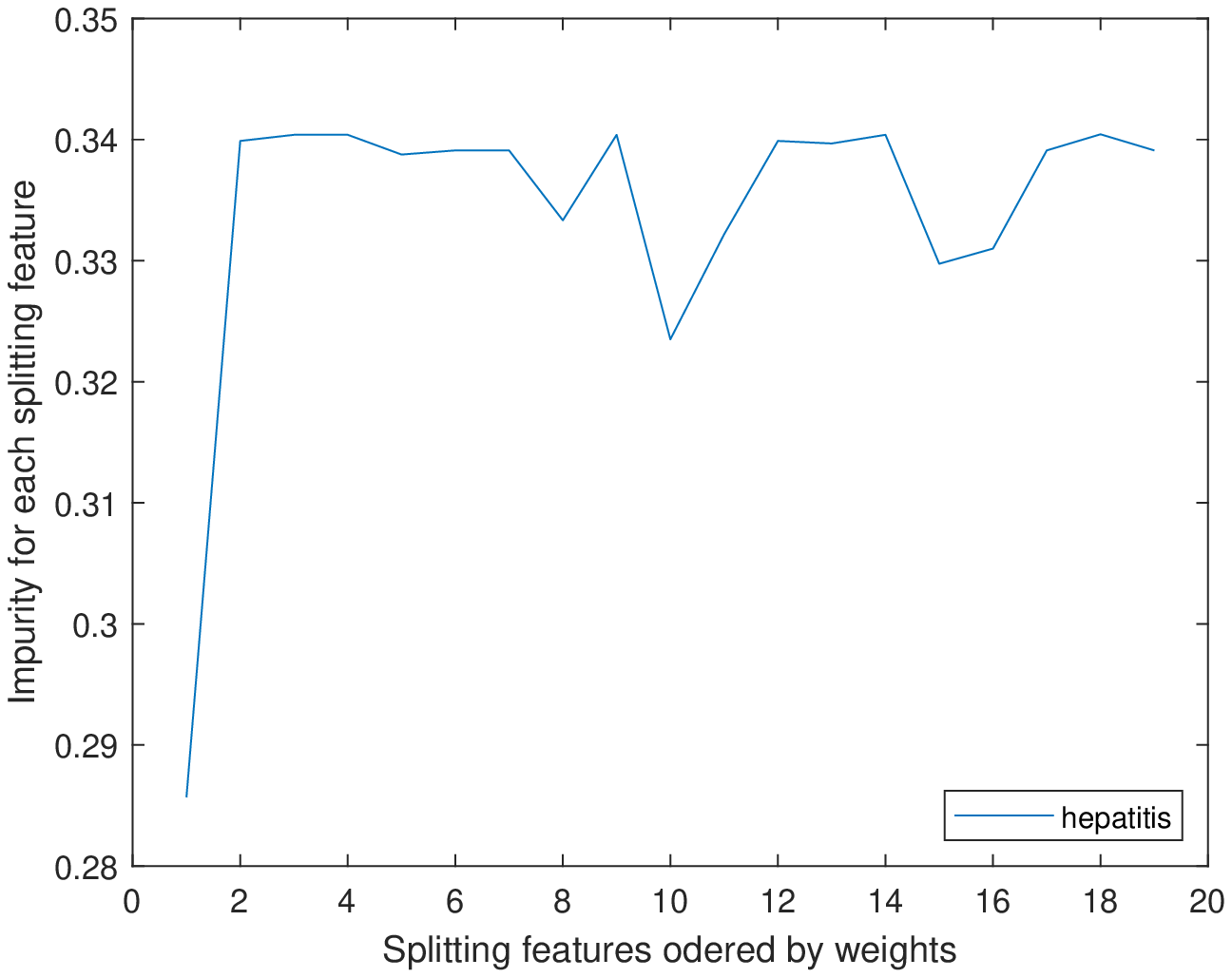}}
    \label{1b}  
    \hfill
   \subfigure[molec biol promoter]{
        \includegraphics[width=0.3\linewidth]{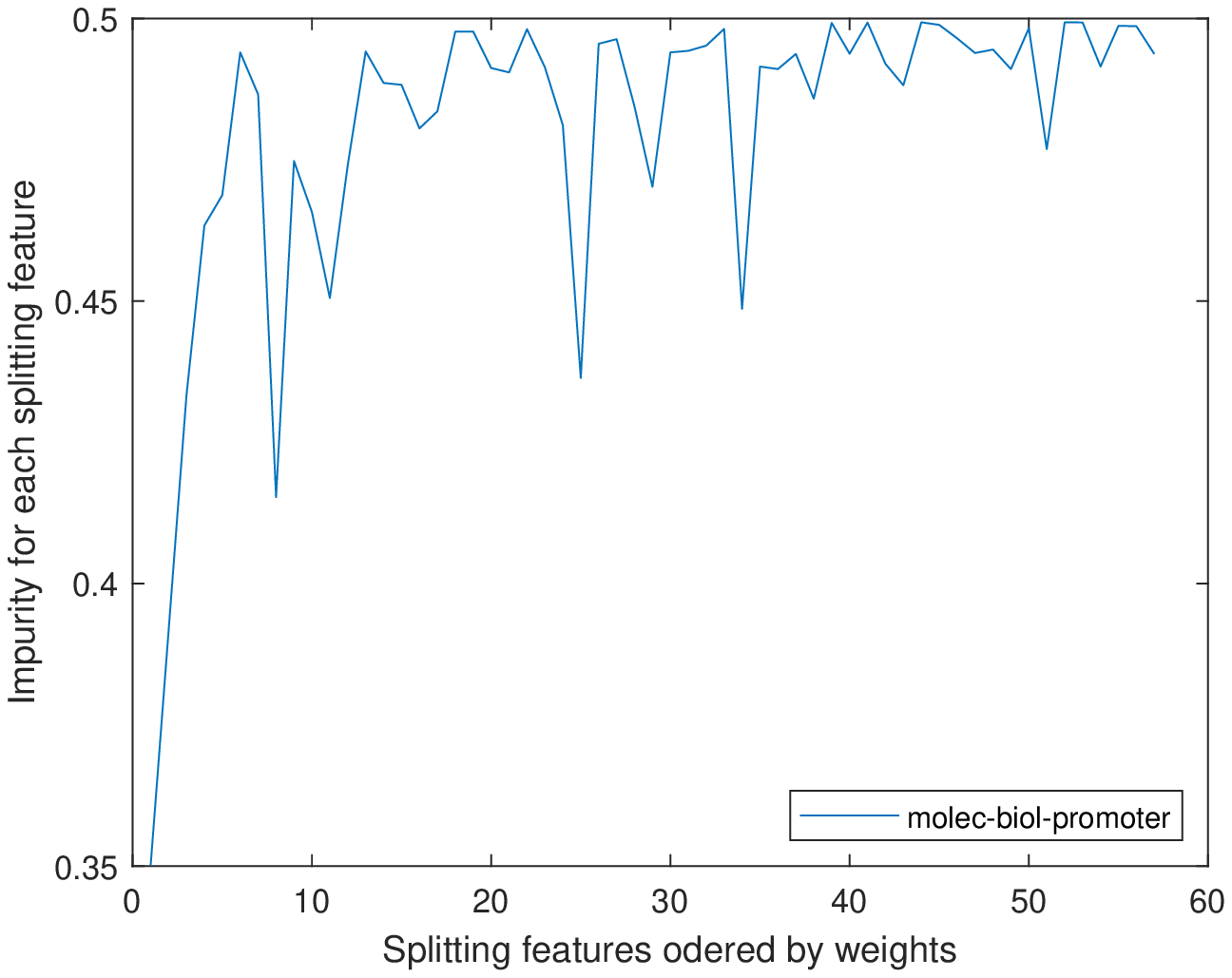}}  \\
    \label{1c} 
   \subfigure[parkinsons]{
        \includegraphics[width=0.3\linewidth]{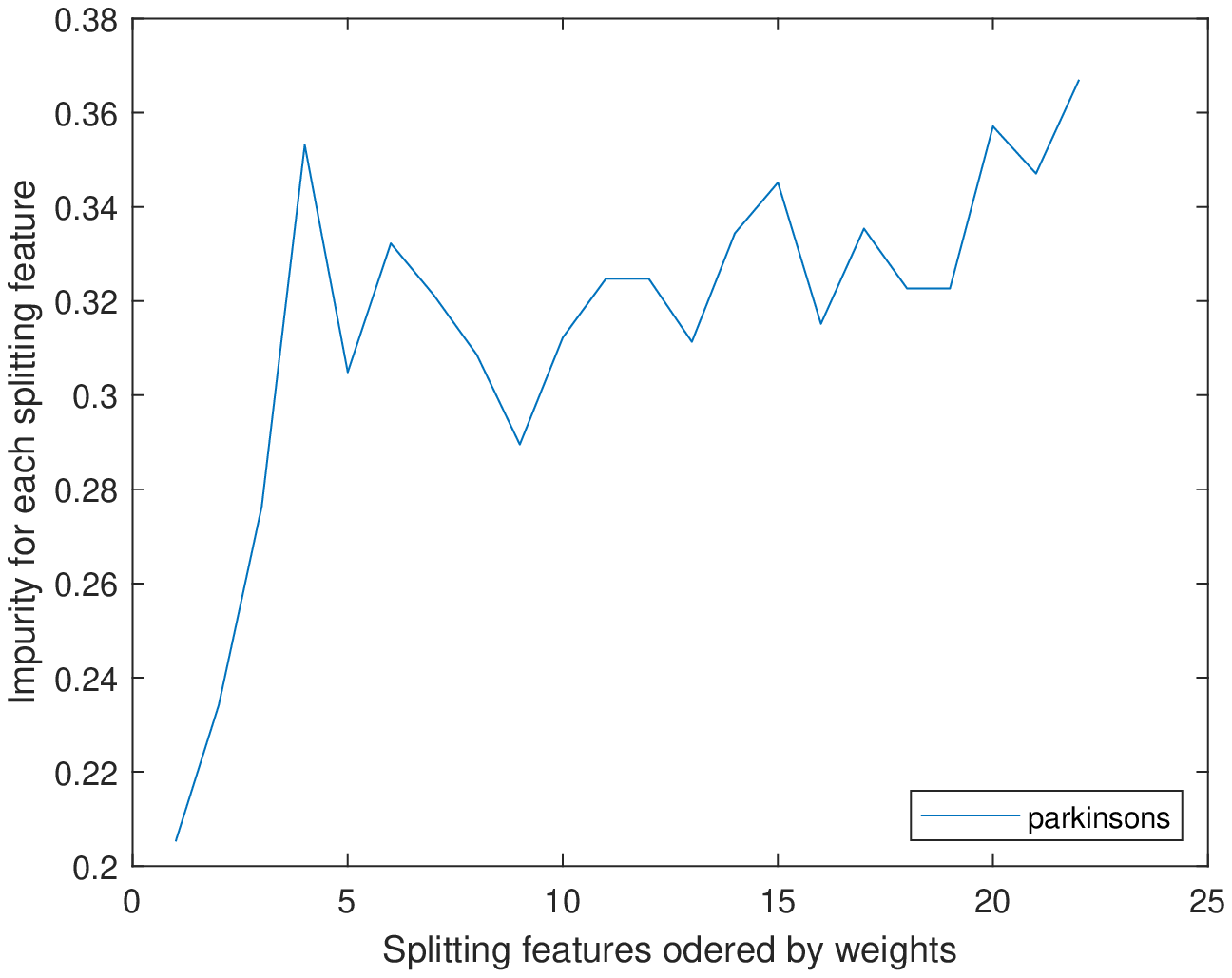}}
     \label{1d}  
     \hfill
   \subfigure[sonar]{
        \includegraphics[width=0.3\linewidth]{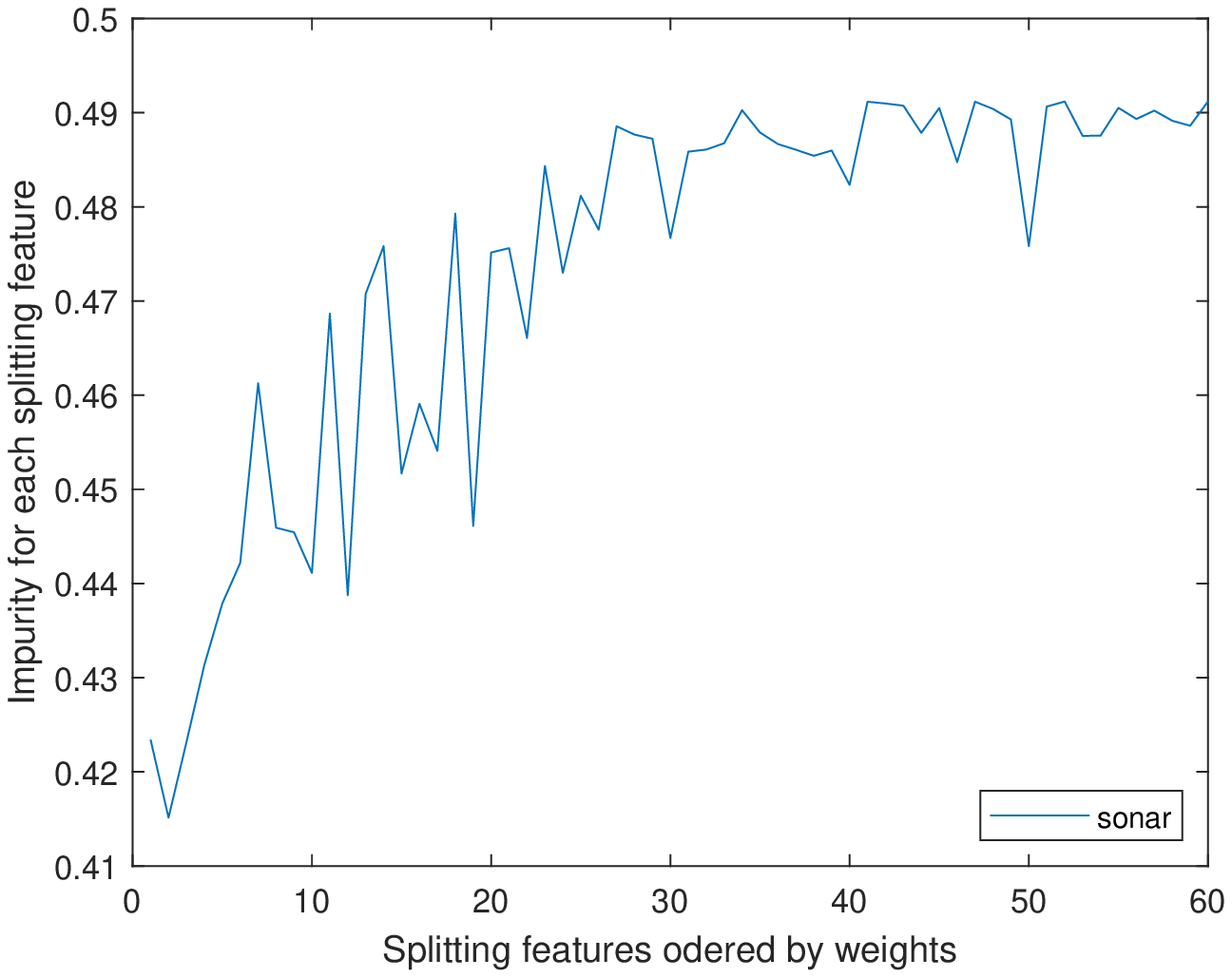}}
     \label{1f} 
     \hfill
   \subfigure[spectf]{
        \includegraphics[width=0.3\linewidth]{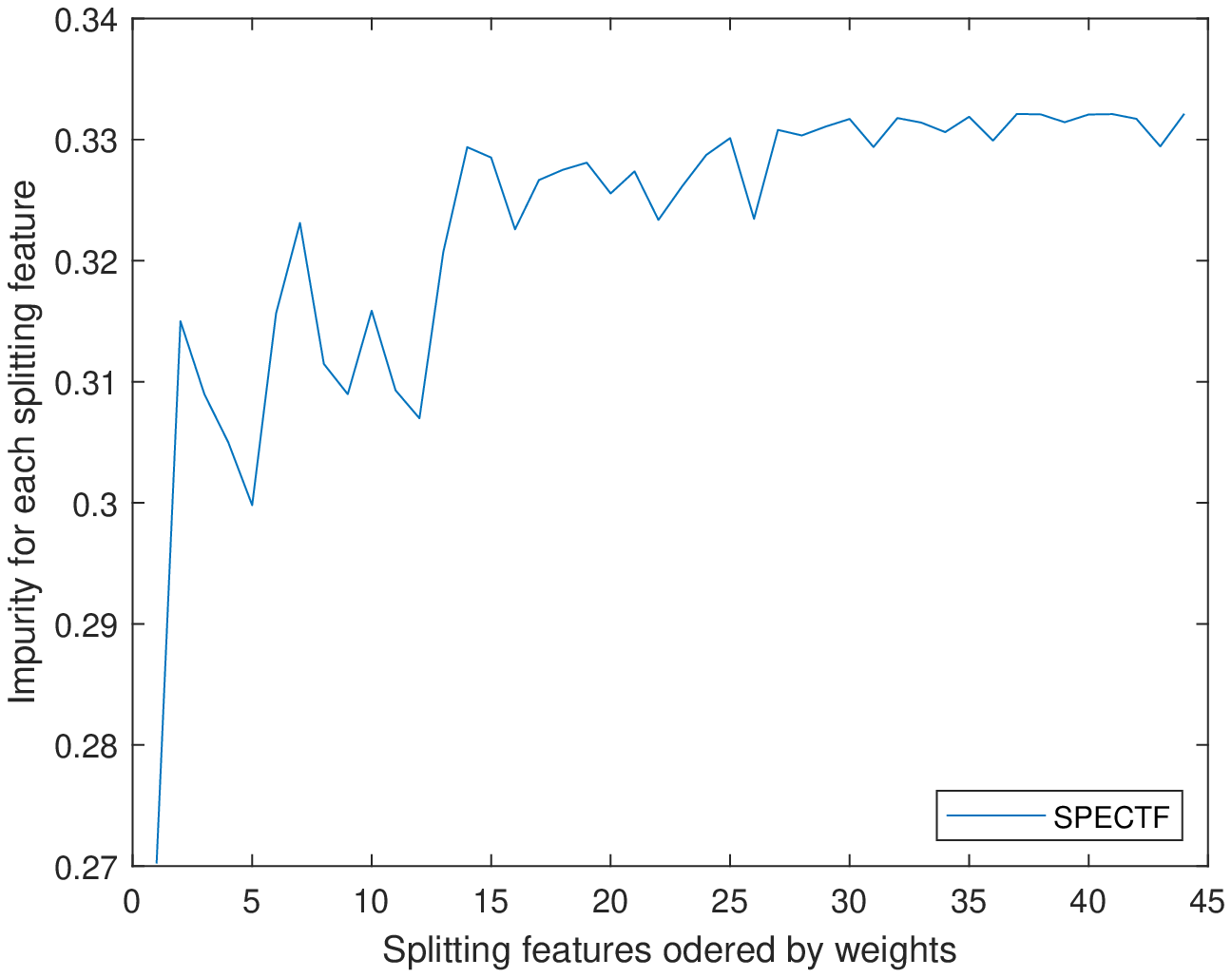}}
     \label{1g} 
   \caption{The relationship between feature importance weights and impurity.}
      \label{featurevsimpurity}
\end{figure*}

\subsection{Inference Runtime}
Firstly, we analyze the computational complexity of DTs theoretically. We are only concerned with a single node rather than the whole tree for simplicity. Suppose there is $n$ samples with $d$ features in a node. In conventional UDTs, such as C4.5 and CART, the complexity is $O(nd \ log(n))$\cite{2015Oblique}, since the best splitting feature and point are selected based on some impurity criteria by traversing the whole feature space. However, for our proposed dGMML-DT, exhaustive search is circumvented, so the computational complexity of it is $O(d \ log(n))$, $n$ times less than conventional UDTs. We compare the training time of dGMML-DT and other DTs, the average training time of each DTs on different datasets is shown in Table \ref{accuracy}. The unit of time is milliseconds and the minimum training time is marked by underline. The inference runtime of dGMML-DT is 5-15 times faster than other existing models. 
\subsection{Feature Weight versus Impurity}
If we choose the most discriminative splitting feature via formula (\ref{10}), is it possible to gain purer children nodes and further build a tree model with better performance? In this subsection, We calculate the importance weight of each feature and rank them in descending order. Subsequently, these ranked features are used to split the dataset respectively and the corresponding impurity is recorded. In Fig. \ref{featurevsimpurity}, x-axis represents the ranked features and y-axis represents the impurity of two partitions after splitting the dataset using the feature. The coordinate where x=1 corresponds to the feature with the largest weight. In general, the impurity of the partitions divided by the feature with larger weight is lower. However, the curve fluctuates locally sometimes, which means that the splitting criterion of dGMML-DT is not completely consistent with splitting criteria based on impurity.

\begin{table}[t]
\centering  
\caption{Comparison of three types of splitting points }  
\label{ablation study}  
\begin{tabular}{|c|c|c|c|}
\cline{1-4}
Formulation   & (\ref{mean12}) & (\ref{medi})  & (\ref{mean})      \\
\cline{1-4}
Average accuracy & 80.7364  & 80.2298  & 79.5642 \\
\cline{1-4}
\end{tabular}
\end{table}

\subsection{Comparison of Three Types of Splitting Points}
In this subsection, we show performance of the three types of splitting strategies (see formula (\ref{mean12})-(\ref{mean})). As represented in table \ref{ablation study}, we report their average prediction accuracies of the same 25 datasets and can find that their differences in accuracy are not significant.

\section{Conclusion} \label{conclusion}
In this paper, we design an efficient and effective splitting criterion to obtain the importance weight of each candidate feature at an internal node. Our method accelerates the tree construction process as it circumvents exhaustive search over the whole feature space. Most importantly, it is flexible enough to be expanded into its multivariable counterpart or embedded into other variations of DTs. Experiments over 25 datasets provide evidence that the performance of dGMML-DT is comparable or superior to other baseline decision trees. Splitting nodes based on some discriminative feature selection methods rather than exhaustive searches is a direction worth exploring. Besides, further research should be carried out to evaluate the discriminative weights of categorical features.

\section*{Acknowledgment}
This work is supported by NSFC under granted No. 62076124.

\bibliographystyle{IEEEtran}
\bibliography{IEEEfull}

\end{document}